\RequirePackage{fix-cm}
\documentclass[twocolumn]{svjour3}          
\smartqed  
\usepackage{graphicx}
\usepackage{graphicx}
\usepackage{multirow}
\usepackage{epsfig}
\usepackage{graphicx}
\usepackage{amsmath}
\usepackage{amssymb}
\usepackage{bm}
\usepackage{mathtools}
\usepackage{algorithm,algcompatible,lipsum}
\usepackage{verbatim}
\usepackage{cite} 
\usepackage{flushend}

\usepackage{array}
\newcolumntype{P}[1]{>{\centering\arraybackslash}p{#1}}

\usepackage[pangram]{blindtext}
\usepackage{soul}
\usepackage{balance}

\def\l{{\ell}}          
\def\R{\mathbb{R}}

\def\E{{\mathbb{E}}} 

\def\batchsize{{|\beta|}}
\def\paramGlobal{{\lambda}}

\begin{document}

\title{
Stochastic batch size for adaptive regularization \\in deep network optimization
}

\titlerunning{Stochastic batch size in deep optimization}        

\author{Kensuke Nakamura \and
        Stefano Soatto \and
        Byung-Woo Hong* 
}


\institute{
K. Nakamura \at Computer Science Department, Chung-Ang University, Seoul, Korea\\
\email{kensuke@image.cau.ac.kr}           
\and
S. Soatto \at Computer Science Department, University of California Los Angeles, CA, U.S.A.\\
\email{soatto@cs.ucla.edu}
\and
B.-W. Hong (*corresponding auther) \at Computer Science Department, Chung-Ang University, Seoul, Korea\\
\email{hong@cau.ac.kr}  
}


\date{Received: date / Accepted: date}

\maketitle

\begin{abstract}
We propose a first-order stochastic optimization algorithm incorporating adaptive regularization applicable to machine learning problems in deep learning framework. The adaptive regularization is imposed by stochastic process in determining batch size for each model parameter at each optimization iteration. The stochastic batch size is determined by the update probability of each parameter following a distribution of gradient norms in consideration of their local and global properties in the neural network architecture where the range of gradient norms may vary within and across layers. We empirically demonstrate the effectiveness of our algorithm using an image classification task based on conventional network models applied to commonly used benchmark datasets. The quantitative evaluation indicates that our algorithm outperforms the state-of-the-art optimization algorithms in generalization while providing less sensitivity to the selection of batch size which often plays a critical role in optimization, thus achieving more robustness to the selection of regularity.
\end{abstract}

%
%
\section{Introduction} \label{sec:introduction}
In deep learning, the inference process is made through a deep network architecture that generally consists of a nested composition of linear and nonlinear functions leading to a large-scale optimization problem where both the number of model parameters and the number of data are in the millions~\cite{bottou2010large,simonyan2014very,szegedy2015going,he2016deep,he2016identity,huang2017densely,bottou2018optimization}.
The considered optimization problems generally aim to minimize non-convex objective functions defined over a large number of training data, which is often computationally challenging.
In dealing with such problems, the most popular algorithm is stochastic gradient descent (SGD)~\cite{robbins1951stochastic,rumelhart1988learning,zhang2004solving,bottou2010large,bottou2018optimization} that randomly selects a subset of training data in evaluating the loss function and its gradient at each iteration while ordinary gradient descent~\cite{ruder2016overview,robbins1951stochastic} uses the entire data.
The estimates of gradients using small batch sizes are generally unreliable, but the computation of full-batch gradients is often intractable, yielding a trade-off between stability and efficiency.
\par
In this work, we propose a stochastic algorithm in the selection of batch size for estimating gradients in the course of optimization in which the batch size is adaptively determined for each parameter at each iteration. The intrinsic motivation stems from a need to impose different degree of regularization on each parameter at each optimization stage in such a way that the generalization is improved by implicitly guiding the trajectory of gradients to preferred minima. 
Our stochastic scheme is designed to determine the batch size of each parameter following a distribution that is proportional to the gradient norm; its gradient with varying batch size is efficiently computed by the cumulative moving average of usual stochastic gradients due to the additive form of the objective function.
The selection of batch size in stochastic optimization is related to regularization in addition to the computational efficiency~\cite{bengio2012practical,li2014efficient} in that the variance of gradients is reduced with larger batch size, allowing one to use a larger learning rate. However, it is not necessarily beneficial for complex non-convex problems in order to avoid undesirable local minima~\cite{Keskar2016OnLT}. 
Conventional SGD uses deterministic selection strategies for the batch size that is mostly static, although dynamic scheduling methods exist~\cite{smith2017don,Liu2018FastVR} without taking into account the relative importance of individual parameter.
\par
In the computation of gradients, SGD uses the same batch to compute gradients of all the parameters while coordinate descent (CD)~\cite{mazumder2011sparsenet,breheny2011coordinate,nesterov2012efficiency} and its block based variants~\cite{richtarik2014iteration,dang2015stochastic,lu2015complexity,zhao2014accelerated} update a subset of parameters at each iteration using the same size of batch for all the parameters. However, the partial update scheme at each batch iteration is inefficient due to the common parallel implementation that can provide gradients of all the parameters at once, resulting in discarding the gradients of non-updated parameters.
\par
We relate our method to prior work in Section~\ref{sec:works} and present classical SGD, CD and their integrated algorithms in Section~\ref{sec:preliminary} followed by our proposed algorithm in Section~\ref{sec:algorithm}. The effectiveness of our algorithm is demonstrated by numerical results in Section~\ref{sec:results} and the conclusion follows in Section~\ref{sec:conculusion}.
%
%
%
\section{Related work} \label{sec:works}
Deep neural networks have made a significant progress in a variety of applications for understanding visual scenes~\cite{long2015fully,Jin2017Inverse,cai2018learning}, sound information~\cite{aytar2016soundnet,Li2017ACO,mesaros2018detection,van2013deep}, physical motions~\cite{dosovitskiy2015flownet,Chen_2015_ICCV,finn2017deep,shahroudy2018deep}, and other decision processes~\cite{owens2018audio,Ahmad2019EventRecog,wang2015collaborative,cheng2016wide}. Their optimization algorithms related to our work are reviewed in the following. 
\par
\noindent {\bf Learning rate:}
In the application of gradient based optimization algorithms, it is generally necessary to determine the step size at updating unknowns. 
In addition to the fixed learning rate, there have been a number of learning rate scheduling schemes used for the SGD in order to achieve better generalization and convergence. 
One of the simple, yet popular schemes is staircase~\cite{smith2017don} and exponential decay~\cite{george2006adaptive} also has been applied to reduce stochastic noises and achieve stable convergence.
The parameter-wise adaptive learning rate scheduling has also been developed such as AdaGrad~\cite{duchi2011adaptive}, AdaDelta~\cite{schaul2013no,zeiler2012adadelta},  RMSprop~\cite{tieleman2012lecture}, and Adam~\cite{kingma2014adam}.
In our experiment we carefully choose the learning rate annealing such that the baseline SGD achieves the best validation accuracy in order to concentrate on the regularization effect due to the batch size.
\par
\vspace{3pt}
\noindent {\bf Variance reduction:} 
The variance of stochastic gradients is detrimental to SGD, motivating variance reduction techniques~\cite{Roux2012,johnson2013accelerating,Chatterji2018OnTT,zhong2014fast,shen2016adaptive,Difan2018svrHMM,Zhou2019ASim} that aim to reduce the variance incurred due to their stochastic process of estimation, and improve the convergence rate mainly for convex optimization while some are extended to non-convex problems~\cite{allen2016variance,huo2017asynchronous,liu2018zeroth}. 
One of the most practical algorithms for better convergence rates includes momentum~\cite{sutton1986two}, modified momentum for accelerated gradient~\cite{nesterov1983method}, and stochastic estimation of accelerated gradient (Accelerated-SGD)~\cite{Kidambi2018Acc}.
These algorithms are more focused on the efficiency in convergence than the generalization of model for accuracy.
\par
\vspace{3pt}
\noindent {\bf Energy landscape:}
Understanding of energy surface geometry is significant in deep optimization of highly complex non-convex problems. 
It is preferred to drive a solution toward plateau in order to yield better generalization~\cite{hochreiter1997flat,Chaudhari2017EntropySGD,dinh2017sharp}.
Entropy-SGD~\cite{Chaudhari2017EntropySGD} is an optimization algorithm biased toward such a wide flat local minimum.
In our approach, we do not attempt to explicitly measure geometric property of loss landscape with extra computational cost, but instead implicitly consider the variance by varying batch size.
\par
\vspace{3pt}
\noindent {\bf Importance sampling} 
aims to limit the estimation of gradients to informative samples for efficiency and variance reduction by considering manual selection~\cite{bengio2009curriculum}, temporal history of gradient~\cite{schaul2015prioritized,loshchilov2015online}, maximizing the diversity of losses~\cite{wu2017sampling,fan2017learning}, or largest changes in parameters~\cite{csiba2015stochastic,konevcny2017semi,Stich2017Approximate,johnson2018training,katharopoulos2018not,borsos2018online}.
Albeit the benefit of non-uniform sampling, it is computationally expensive to estimate distribution for importance of huge number of parameters. 
\par
\vspace{3pt}
\noindent {\bf Dropout}
is an effective regularization technique in particular with shallower networks by ignoring randomly selected units following a certain probability during the training phase~\cite{srivastava2014dropout}. 
The dropping rate is generally set to be constant, typically $0.5$, but its variants have been proposed with adaptive rates depending on parameter value~\cite{ba2013adaptive}, estimated gradient variance~\cite{kingma2015variational}, biased gradient estimator~\cite{srinivas2016generalized}, layer depth~\cite{huang2016deep}, or marginal likelihood over noises~\cite{noh2017regularizing}.
\par
\vspace{3pt}
\noindent {\bf Coordinate descent and its variants:}
In contrast to SGD that selects a random subset of data for updating all the parameters at each iteration, coordinate descent (CD) updates a random subset of parameters using all the data~\cite{warga1963minimizing,bertsekas1989parallel}.
It has been proposed to integrate SGD and CD resulting in stochastic random block coordinate descent~\cite{richtarik2014iteration,zhao2014accelerated} that selects a random subset of data to update a random subset of parameters.
In selection of parameters to update, greedy coordinate descent~\cite{Qi2016cd,song2017accelerated,Diakonikolas2018AlternatingRB} and its stochastic approach~\cite{richtarik2014iteration,noh2017regularizing,nutini2017let} are designed to follow a distribution of gradient norms.
Our algorithm is developed in the framework of stochastic greedy block coordinate descent incorporating stochastic selection of batch size.
\par
\vspace{3pt}
\noindent {\bf Batch size selection:}
There is a trade-off between the computational efficiency and the stability of gradient estimation leading to the selection of their compromise with generally a constant while learning rate is scheduled to decrease for convergence.
The generalization effect of stochastic gradient methods has been analyzed with constant batch size~\cite{Moritz2016train,he2019control}.
On the other hand, increasing batch size per iteration with a fixed learning rate has been proposed in~\cite{smith2017don} where the equivalence of increasing batch size to learning rate decay is demonstrated.
A variety of varying batch size algorithms have been proposed by variance of gradients~\cite{balles2016coupling,levy2017online,de2017automated,Liu2018FastVR} where the additional computational cost for the gradient variance is expensive and the same batch size is used at each estimation of gradients for all the parameters.
%
%
%
%
%
%
%
%
%
\section{Preliminaries} \label{sec:preliminary}
We consider a minimization problem of an objective function $F \colon \R^m \to \R$ in a supervised learning framework:
\begin{align}
    w^* = \arg\min_w F(w), \label{eq:minimization}
\end{align}
%
where $F$ is associated with $w = (w_1, w_2, \cdots, w_m)$ in the finite-sum form:
\begin{align}
    F(w) = \frac{1}{n} \sum_{i=1}^n \l(h_w(x_i), y_i) = \frac{1}{n} \sum_{i=1}^n f_i(w), \label{eq:energy}
\end{align}
%
%
where $h_w \colon X \to Y$ is a prediction function defined with the associated model parameters $w$ from a data space $X$ to a label space $Y$, and $f_i(w) \coloneqq \l(h_w(x_i), y_i)$ is a differentiable loss function defined by the discrepancy between the prediction $h_w(x_i)$ and the true label $y_i$. 
The objective is to find optimal parameters $w^*$ by minimizing the empirical loss incurred on a given set of training data $\{(x_1, y_1), (x_2, y_2), \cdots, (x_n, y_n)\}$.
%
%
\subsection{Stochastic gradient descent} \label{sec:sgd}
The optimization of supervised learning applications that often require a large number of training data mainly uses stochastic gradient descent that updates solution $w^{t}$ at each iteration $t$ based on the gradient:
\begin{align} 
    w^{t+1} = w^{t} - \eta^t \left( \nabla F(w^{t}) + \xi^t \right), \label{eq:update.sgd}
\end{align}
%
%
%
where $\eta^t \in \R$ is a learning rate and $\xi^t$ is an independent noise process with zero mean.
The computation of gradient for the entire training data is computationally expensive and often intractable so that stochastic gradient is computed using batch $\beta^t$ at each iteration $t$:
\begin{align}
     w^{t+1} = w^{t} - \eta^t \left( \frac{1}{|\beta^t|} \sum_{i \in \beta^t} \nabla f_i(w^{t}) \right), \label{eq:sgd}
\end{align}
%
where $\beta^t$ is a subset of the index set $[n] = \{1, 2, \cdots, n\}$ for the training data.
The selection of batch size $|\beta^t|$ that is inversely proportional to the variance of $\xi^t$ is often critical.
%
%
\subsection{Alternating minimization} \label{sec:am}
A class of alternating minimization is considered as an effective optimization algorithm to deal with a large number of parameters. 
The alternating minimization method with a cyclic constraint randomly partitions the unknown parameters $w$ into two mutually disjoint subsets $u$ and $v$, and the optimization alternatively proceeds over one block and the other at each iteration $t$:
%
\begin{align}
    u^{t+1} &= u^{t} - \eta^t \, \nabla_u F(u^{t}, v^{t}),   \label{eq:u}\\
    v^{t+1} &= v^{t} - \eta^t \, \nabla_v F(u^{t+1}, v^{t}), \label{eq:v}
\end{align}
where $\nabla_u F$ and $\nabla_v F$ denote the gradient of $F$ with respect to $u$ and $v$ using the entire training data. 
The above alternating minimization algorithm is equivalent to randomized cyclic block coordinate descent with two blocks.
%
%
\subsection{Stochastic alternating minimization} \label{sec:sam}
The optimization of large scale learning problems is often required to deal with a large number of both model parameters and training data, which naturally leads to consideration of combining the use of stochastic gradients based on batches and alternating minimization over parameter blocks.  
The combination of stochastic gradient descent and alternating minimization leads to stochastic randomized cyclic block coordinate descent with two blocks where a random subset of parameters are updated based on their stochastic gradients.
The stochastic alternating minimization method finds a solution based on the gradient with respect to each parameter block using batches selected from the training data uniformly at random. 
The algorithm updates blocks $u^{t}$ and $v^{t}$ of parameters $w^{t}$ in an alternative way:
\begin{align}
    u^{t+1} &= u^{t} - \frac{\eta^t}{|\beta^t|} \sum_{i \in \beta^t} \nabla_u f_i(u^{t}, v^{t}), \label{eq:su}\\
    v^{t+1} &= v^{t} - \frac{\eta^t}{|\beta^{t+1}|} \sum_{i \in \beta^{t+1}} \nabla_v f_i(u^{t+1}, v^{t}), \label{eq:sv}
\end{align}
%
where parameter blocks $u$ and $v$ are updated based on the gradients computed using $\beta^t$ and $\beta^{t+1}$, respectively.
%
%
%
%
%
\section{Proposed algorithm} \label{sec:algorithm}
We propose a first-order optimization algorithm that updates each model parameter based on its stochastic gradient computed with stochastic batch size. The batch size for each parameter is determined by its update probability following a distribution of the norm of stochastic gradients in consideration of local and global properties of gradient norms in the network architecture.
%
%
%
\subsection{Alternating minimization with stochastic batch size} \label{sec:amsbs}
The update probability $p_j$ associated with  parameter $w_j$ is represented by a Bernoulli random variable $\chi_j$:
\begin{align}
    \Pr(\chi_j = 1) &= p_j,
\end{align}
%
%
%
where $0 < p_j < 1$ is the probability that $\chi_j = 1$.
The proposed algorithm is designed to update each parameter $w_j$ according to the probability $p_j$ at iteration $t$:
\begin{align}
    w_j^{t+1} &= w_j^{t} - \chi_j^t ( \eta_j^t \, g_j^t ),
\end{align}
where $\eta_j^t$ is a learning rate to update $w_j^t$, 
$\chi_j^t$ is an indicator variable associated with the update probability $p_j^t$ for $w_j^t$, 
and $g_j^t$ is stochastic gradient with respect to $w_j$:
\begin{align}
    g_j^t = \frac{1}{| \beta_j^t |} \sum_{i \in \beta_j^t} \nabla_j f_i(w^t),
\end{align}
where $\nabla_j f = e_j^T \nabla f$ denotes the gradient of $f$ with respect to $w_j$, $e_j$ is a unit vector, and $\beta_j^t$ denotes a batch.
The essence of the proposed algorithm is to determine the size of batch $\beta_j^t$ for each parameter $w_j$ at iteration $t$ depending on the update probability $p_j^t$.
Let $\{\beta^t\}$ be a set of usual batches, called universal batches, that are the same as the ones used by standard SGD.
The stochastic batch $\beta_j^t$ is determined in a recursive manner depending on the update probability $p_j^t$:  
\begin{align}
    \beta_j^t = 
    \begin{cases}
        \beta_j^{t-1} \cup \beta^t, & \text{if } \chi_j^t = 0\\
        \varnothing, & \text{otherwise}
    \end{cases}
\end{align}
where $\beta_j^t$ is obtained by the accumulation of the previous batch $\beta_j^{t-1}$ with the current universal batch $\beta^t$ when parameter $w_j$ is not updated.
On the other hand, $\beta_j^t$ is set to be an empty set after parameter $w_j$ is updated.
The proposed algorithm with constant update probability $p_j^t = 1$ for all $j$ and $t$ results in standard SGD, and $p_j^t = 1/2$ results in an algorithm similar to Dropout except its forward-propagation that assigns zero to the output of ignored node while the proposed algorithm preserves the previous value.
The pseudocode of the algorithm is described in Algorithm~\ref{alg:sbs} where the indicator variable $\chi_j^t$ is determined by the update probability $p_j^t$ that will be presented in Section~\ref{sec:adaptive.update.probability}.
%
%
%
%
%
\begin{algorithm}[htb]
 	\caption{Alternating Minimization with Stochastic Batch Size}
\label{alg:sbs}
\begin{algorithmic}
\normalsize
	\FORALL {epoch}
        \STATE $\{ \beta^t \}$ : universal batches given by random shuffling
        \STATE $\beta_j^0 = \varnothing$ : initialize stochastic batch for all $j$ 
        \FORALL {$t$ : index for universal batch} 
            \FORALL {$j$ : index for parameter}
                \STATE $\beta_j^t = \beta_j^{t-1} \cup \beta^t$
                \IF {$\chi_j^t = 1$}
                    \STATE $g_j^t = \frac{1}{|\beta_j^t|} \sum_{i \in \beta_j^t} \nabla_j f_i(w^t)$
                    \STATE $w_j^{t+1} = w_j^t - \chi_j^t (\eta_j^t \, g_j^t)$
                    \STATE $\beta_j^t = \varnothing$
                \ENDIF
            \ENDFOR
        \ENDFOR
    \ENDFOR
\end{algorithmic}
\end{algorithm}
%
%
%
%
%
%
\subsection{Efficient algorithm of stochastic batch size via cumulative moving average}
The computation of gradient using stochastic batch with varying size is inefficient due to the parallel implementation of back-propagation applied to the mixture of updating and non-updating parameters.
Thus, we propose an alternative efficient algorithm that manipulates the gradients computed using universal batches instead of directly computing gradients using stochastic batches due to the additive form of the objective function. 
The modified algorithm computes gradients of all the parameters using universal batch at each iteration and then takes weighted average of the non-updated previous gradients of each parameter to estimate its gradient with accumulated batches when updating, which leads to a Gauss-Seidel type iterative method.
\par
Let us denote by $\beta^{[k]} = \beta^1 \cup \beta^2 \cup \cdots \cup \beta^k$ a union of universal batches where we assume that $|\beta^1| = |\beta^2| = \cdots = |\beta^k|$ for ease of presentation.
The update of parameter $w_j$ using stochastic batch $\beta^{[k]}$ at iteration $t$ reads:
\begin{align}
w_j^{t+1} = w_j^t - \tilde{\eta}_j^t \left( \frac{1}{|\beta^{[k]}|} \sum_{i \in \beta^{[k]}} \nabla_j f_i(w^t) \right), \label{eq:our_update}
\end{align}
where $\tilde{\eta}_j^t$ denotes a learning rate associated with a stochastic batch $\beta^{[k]}$.
We can consider $\tilde{\eta}_j^t := k \, \eta_j^t$ where $\eta_j^t$ denotes the learning rate for a single universal batch, e.g. $\beta^1$, due to the linear relation between the learning rate and the batch size, and $|\beta^{[k]}| = k |\beta^1|$.
The update in Eq.(\ref{eq:our_update}) requires separate back-propagations for different parameters when they have different stochastic batches.
Therefore we estimate gradient $\tilde{g}_j$ of $w_j$ for a set of universal batches $\beta^{[k]}$ by taking the cumulative moving average of the gradients computed using a series of universal batches leading to the following recursive steps with the initial condition $\tilde{g}_j^t = 0$:
\begin{align} 
    \begin{dcases}
        g_j^{t + \frac{s}{k}} &= \frac{1}{|\beta^s|} \sum_{i \in \beta^s} \nabla_j f_i(w^{t + \frac{s-1}{k}}),\\
        \tilde{g}_j^{t + \frac{s}{k}} &= \tilde{g}_j^{t + \frac{s-1}{k}} + \frac{1}{s} \left(g_j^{t + \frac{s}{k}} - \tilde{g}_j^{t + \frac{s-1}{k}}\right),
    \end{dcases}
\end{align}
where $s$ iterates from $1$ to $k$, and the update of parameter $w_j$ is obtained by:
\begin{align} 
    w_j^{t+1} &= w_j^{t} - \tilde{\eta}_j^t \, \tilde{g}_j^{t+1},
\end{align}
where $\tilde{g}_j^{t+1}$ is considered as gradient of $w_j$ with stochastic batch $\beta^{[k]}$, and $\tilde{\eta}_j^t$ denotes its learning rate. 
This Gauss-Seidel type of recursive update enables the parallelization in computing the gradients with stochastic batches. In addition, the use of intermediate update of other parameters in computing stochastic gradient is beneficial in convergence.
\par
The pseudocode of the modified algorithm is described in Algorithm~\ref{alg:sbs:modified} where the computation of gradients with respect to all the parameters is parallelized at each universal batch iteration, and their cumulative moving averages are used to estimate the gradients with stochastic batches.
Note that the back-propagation cost of the modified algorithm is the same as the conventional SGD using the universal batch. 
%
%
%
%
%
\begin{algorithm}[htb]
	\caption{Alternating minimization with stochastic batch size via cumulative moving average}
\label{alg:sbs:modified}
\begin{algorithmic}
\normalsize
	\FORALL {epoch}
        \STATE $\{ \beta^t \}$ : universal batches given by random shuffling
        \STATE $\tilde{g}_j^0 = 0$ : initialize gradients for all $j$
        \STATE $k_j = 1$ : initialize the count of batches for all $j$
        \FORALL {$t$ : index for universal batch} 
            \STATE $g^t = \frac{1}{|\beta^t|} \sum_{i \in \beta^t} \nabla f_i(w^t)$
            \FORALL {$j$ : index for parameter}
                \STATE $\tilde{g}_j^t = \tilde{g}_j^{t-1} + \frac{1}{k_j} \left( g_j^t - \tilde{g}_j^{t-1} \right)$
                \STATE $w_j^{t+1} = w_j^t - \chi_j^t (\eta_j^t \, \tilde{g}_j^t)$
                \STATE $\tilde{g}_j^t = (1 - \chi_j^t) \, \tilde{g}_j^t$
                \STATE $k_j = (1 - \chi_j^t) (k_j + 1) + \chi_j^t$
            \ENDFOR
        \ENDFOR
    \ENDFOR
\end{algorithmic}
\end{algorithm}
%
%
%
%
%
%
\subsection{Adaptive update probability} \label{sec:adaptive.update.probability}
The update of parameters $w_j$ at iteration $t$ is determined by the associated probability $p_j^t$ based on the sigmoid function: 
\begin{align}
    s(x; \alpha, \gamma) &= \frac{1}{1 + \exp(- \alpha x - \gamma)}, \label{eq:sigmoid}
\end{align}
where $\alpha, \gamma \in \R$ are parameters for the slope and the horizontal position of the curve, respectively.
The update probability is designed to be proportional to the norm of the gradient following the sigmoid function where $\alpha$ and $\gamma$ are related to the randomness and the expectation of the probability, respectively.
We present local and global adaptive schemes in the determination of the update probability and we will use their combination.
In consideration of local and global adaptive probability in a neural network architecture, we denote by $\theta_l$ a set of parameter indexes and $\{ w_j \mid j \in \theta_l \}$ a set of parameters at layer $l$ where $l = 1, 2, \cdots, L$.
%
%
%
%
\par
\vspace{5pt}
\noindent {\bf Local adaptive probability:}
The local scheme is designed to consider the relative magnitude of the gradient norms in determining the update probability of the parameters within each layer.
Let $\mu_l^t$ and $\sigma_l^t$ be the mean and the standard deviation (std) of each set of gradient norms $\{| g_j^t |\}_{j \in \theta_l}$ of parameters $\{w_j\}_{j \in \theta_l}$ at iteration $t$. 
Note that we compute the weighted average considering the number of parameters at different layers. 
We also consider the parameter types such as convolution, fully connected, bias, batch normalization in constructing distribution of gradient norms since different types of parameters tend to have different ranges of gradient norms.
We initially normalize the gradient norms of parameters at each layer with mean $0$ and std $1$:
\begin{align}
    v_j^t &= \frac{| g_j^t | - \mu_l^t}{\sigma_l^t}, \quad \forall j \in \theta_l, \quad l = 1, 2, \cdots, L. \label{eq:white:grad}
\end{align}
%
Then, the update probability $p_j^t$ is determined by:
\begin{align}
    p_j^t = s(v_j^t; \alpha, \gamma), \quad \gamma = 0, \label{eq:prob:local}
\end{align}
where $\alpha \in \R$ is a control parameter for the randomness of update, and $\alpha = 0, \gamma = 0$ yields a random update with probability $1/2$.
Setting $\gamma = 0$ leads to $\E_t \left[ p_j^t \mid j \in \theta_l \right] = 1/2$ for all $l$, thus we equally consider all the layers in updating their parameters.
%
%
%
\par
\vspace{5pt}
\noindent {\bf Global adaptive probability:}
The global scheme is designed to consider the relative magnitude of the average gradient norms from all the layers, and assign the same probability to all the parameters within the same layer.
Let $\mu^t$ and $\sigma^t$ be the mean and the std of $\mu_l^t$ over all the layers $l = 1, 2, \cdots, L$ at iteration $t$. 
We normalize the mean $\mu_l^t$ over layers $l$ with mean $0$ and std $1$:
\begin{align}
    \tilde{\mu}_l^t &= \frac{\mu_l^t - \mu^t}{\sigma^t}, \quad l = 1, 2, \cdots, L. \label{eq:white:mean}
\end{align}
%
Then, the update probability $p_j^t$ for all the parameters $\{w_j\}_{j \in \theta_l}$ at layer $l$ is determined by:
\begin{align}
    p_j^t = s(v_j^t; \alpha, \gamma), \quad \alpha = 0, \quad \gamma = \lambda \, \tilde{\mu}_l^t, \quad \forall j \in \theta_l \label{eq:prob:global}
\end{align}
where $\lambda \in \R$ is a control parameter for the randomness of update across layers and the gradient norm of individual parameters is ignored by setting $\alpha = 0$ leading to $\E_l\left[ p_j^t \mid j \in \theta_l \right] = 1/2$. The same update probability is applied to all the parameters at each layer, but the probability is adaptively determined across layers depending on the mean of the gradient norms at each layer. The global scheme effectively considers different ranges of gradient norms at different layers.
%
%
%
%
%
%
%
%
%
%
%
%
%
\par
\vspace{5pt}
\noindent {\bf Combined adaptive probability:}
Our final choice of the update probability uses both local and global adaptive schemes considering relative magnitude of gradient norms of parameters both within- and cross-layers. 
The gradient norm of each parameter is normalized at each layer and the weighted means of the gradient norms with the number of parameters are normalized across layers. 
The combined update probability $p_j^t$ is determined by:
\begin{align}
    p_j^t = s(v_j^t; \alpha, \gamma), \quad \gamma = \lambda \, \tilde{\mu}_l^t, \quad \forall j \in \theta_l \label{eq:prob:combine}
\end{align}
where $\alpha, \paramGlobal \in \R$ are constants related to the std of the update probabilities within- and cross-layers, respectively.
%
%
%
%
%
%
%
%
%
%

%
%
\begin{table*}[htb]
\caption{Validation accuracy ($\%$) of SGD using VGG11 (left) and ResNet18 (right) with different learning rate annealing schemes: constants ($0.1, 0.01, 0.001$), Exponential decay~\cite{george2006adaptive}, RMSProp~\cite{tieleman2012lecture}, Adam~\cite{kingma2014adam}, Staircase~\cite{smith2017don}, and our Sigmoid based on CIFAR10 with batch size $|\beta| = 16$ and momentum = 0. The average accuracy (upper) over the last 10$\%$ of epochs and the maximum (lower) are computed over 10 trials.}
\label{tab:LRschedules.accuracy}
\setlength\extrarowheight{3pt}
\def \pw {5.5mm}
\def \qw {8mm}
\centering
\small
\begin{tabular}{l | P{\pw}P{\pw}P{\pw}P{\pw}P{\pw}P{\pw}P{\pw}P{\qw} | P{\pw}P{\pw}P{\pw}P{\pw}P{\pw}P{\pw}P{\pw}P{\qw}}
\hline
 & \multicolumn{8}{c|}{VGG11}  &  \multicolumn{8}{c}{ResNet18}  \\
\cline{2-17}
& 0.1 & 0.01 & 0.001 & Exp & RMS & Adam & Stair & Sigm & 0.1 & 0.01 & 0.001 & Exp & RMS & Adam & Stair & Sigm\\
\hline
mean & 83.44 & 88.43 & 87.36 & 91.23 & 88.55 & 88.58 & 91.88 & \textbf{92.03} & 87.76 & 91.45 & 90.23 & 93.78 & 91.11 & 91.35 & 94.69 & \textbf{94.89}\\
max & 86.31 & 89.56 & 88.04 & 91.98 & 89.42 & 89.30 & 92.37 & \textbf{92.48} & 90.33 & 92.66 & 91.01 & 94.59 & 91.88 & 92.26 & 95.10 & \textbf{95.16}\\
\hline
\end{tabular}
\vspace{3pt}
\end{table*}
%
\graphicspath{{./figure/img/}}
\def\pw{50pt}  
\def\fwa{50pt} 
\def\fw{60pt}  
\begin{figure*} [htb]
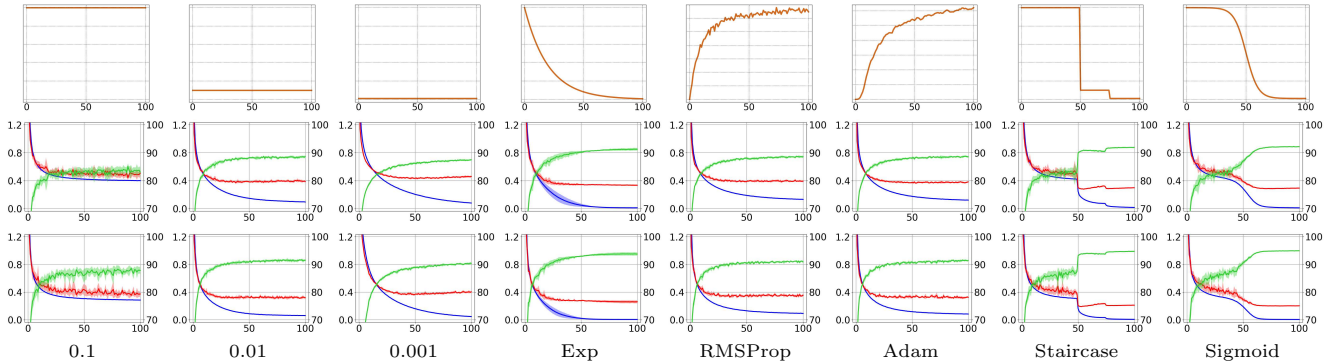

\scriptsize
\centering
%
\begin{tabular}{P{\pw}P{\pw}P{\pw}P{\pw}P{\pw}P{\pw}P{\pw}P{\pw}}
\hspace*{4.5pt} \includegraphics[width=\fwa]{{{lr_cifar10,VGG11,e,100,B,16,nB,1,dB,1,M,1,fixed0.1,1.0,0.0,0.0,0.0,mo,0.0,tr,1,fixed0.1}}}  &
\hspace*{4.5pt} \includegraphics[width=\fwa]{{{lr_cifar10,VGG11,e,100,B,16,nB,1,dB,1,M,1,fixed0.01,1.0,0.0,0.0,0.0,mo,0.0,tr,1,fixed0.01}}}  &
\hspace*{4.5pt} \includegraphics[width=\fwa]{{{lr_cifar10,VGG11,e,100,B,16,nB,1,dB,1,M,1,fixed0.001,1.0,0.0,0.0,0.0,mo,0.0,tr,1,fixed0.001}}}  &
\hspace*{4.5pt} \includegraphics[width=\fwa]{{{lr_cifar10,VGG11,e,100,B,16,nB,1,dB,1,M,1,exp20,1.0,0.0,0.0,0.0,mo,0.0,tr,1,exp20}}}  &
\hspace*{4.5pt} \includegraphics[width=\fwa]{{{lr_cifar10,VGG11,e,100,B,16,nB,1,dB,1,M,1,RMSprop,0.95,0.0001,0.0,0.0,mo,0.0,tr,1,RMSprop}}}  &
\hspace*{4.5pt} \includegraphics[width=\fwa]{{{lr_cifar10,VGG11,e,100,B,16,nB,1,dB,1,M,1,Adam,0.0001,0.9,0.999,0.0,mo,0.9,tr,1,Adam}}}  &
\hspace*{4.5pt} \includegraphics[width=\fwa]{{{lr_cifar10,VGG11,e,100,B,16,nB,1,dB,1,M,1,step,1.0,0.0,0.0,0.0,mo,0.0,tr,1,step}}}  &
\hspace*{4.5pt} \includegraphics[width=\fwa]{{{lr_cifar10,VGG11,e,100,B,16,nB,1,dB,1,M,1,sigmoid,1.0,0.0,0.0,0.0,mo,0.0,tr,1,sigmoid}}}  \\
\includegraphics[width=\fw]{{{ep-ac,cifar10,VGG11,e,100,B,16,nB,1,dB,1,M,1,fixed0.1,1.0,0.0,0.0,0.0,mo,0.0,fixed0.1}}}  &
\includegraphics[width=\fw]{{{ep-ac,cifar10,VGG11,e,100,B,16,nB,1,dB,1,M,1,fixed0.01,1.0,0.0,0.0,0.0,mo,0.0,fixed0.01}}}  &
\includegraphics[width=\fw]{{{ep-ac,cifar10,VGG11,e,100,B,16,nB,1,dB,1,M,1,fixed0.001,1.0,0.0,0.0,0.0,mo,0.0,fixed0.001}}}  &
\includegraphics[width=\fw]{{{ep-ac,cifar10,VGG11,e,100,B,16,nB,1,dB,1,M,1,exp20,1.0,0.0,0.0,0.0,mo,0.0,exp20}}}  &
\includegraphics[width=\fw]{{{ep-ac,cifar10,VGG11,e,100,B,16,nB,1,dB,1,M,1,RMSprop,0.95,0.0001,0.0,0.0,mo,0.0,RMSprop}}}  &
\includegraphics[width=\fw]{{{ep-ac,cifar10,VGG11,e,100,B,16,nB,1,dB,1,M,1,Adam,0.0001,0.9,0.999,0.0,mo,0.9,Adam}}}  &
\includegraphics[width=\fw]{{{ep-ac,cifar10,VGG11,e,100,B,16,nB,1,dB,1,M,1,step,1.0,0.0,0.0,0.0,mo,0.0,step}}}  &
\includegraphics[width=\fw]{{{ep-ac,cifar10,VGG11,e,100,B,16,nB,1,dB,1,M,1,sigmoid,1.0,0.0,0.0,0.0,mo,0.0,sigmoid}}} \\

\includegraphics[width=\fw]{{{ep-ac,cifar10,ResNet18,e,100,B,16,nB,1,dB,1,M,1,fixed0.1,1.0,0.0,0.0,0.0,mo,0.0,fixed0.1}}}  &
\includegraphics[width=\fw]{{{ep-ac,cifar10,ResNet18,e,100,B,16,nB,1,dB,1,M,1,fixed0.01,1.0,0.0,0.0,0.0,mo,0.0,fixed0.01}}}  &
\includegraphics[width=\fw]{{{ep-ac,cifar10,ResNet18,e,100,B,16,nB,1,dB,1,M,1,fixed0.001,1.0,0.0,0.0,0.0,mo,0.0,fixed0.001}}}  &
\includegraphics[width=\fw]{{{ep-ac,cifar10,ResNet18,e,100,B,16,nB,1,dB,1,M,1,exp20,1.0,0.0,0.0,0.0,mo,0.0,exp20}}}  &
\includegraphics[width=\fw]{{{ep-ac,cifar10,ResNet18,e,100,B,16,nB,1,dB,1,M,1,RMSprop,0.95,0.0001,0.0,0.0,mo,0.0,RMSprop}}}  &
\includegraphics[width=\fw]{{{ep-ac,cifar10,ResNet18,e,100,B,16,nB,1,dB,1,M,1,Adam,0.0001,0.9,0.999,0.0,mo,0.9,Adam}}}  &
\includegraphics[width=\fw]{{{ep-ac,cifar10,ResNet18,e,100,B,16,nB,1,dB,1,M,1,step,1.0,0.0,0.0,0.0,mo,0.0,step}}}  &
\includegraphics[width=\fw]{{{ep-ac,cifar10,ResNet18,e,100,B,16,nB,1,dB,1,M,1,sigmoid,1.0,0.0,0.0,0.0,mo,0.0,sigmoid}}} \\
\quad 0.1 &
\quad 0.01 &
\quad 0.001 &
\quad Exp &
\quad RMSProp &
\quad Adam &
\quad Staircase &
\quad Sigmoid\\
\end{tabular}
%
%
\caption{Learning rate annealing curves (top) given by the constants ($0.1$, $0.01$, $0.01$), Exponential decay~\cite{george2006adaptive}, RMSProp~\cite{tieleman2012lecture}, Adam~\cite{kingma2014adam}, Staircase~\cite{smith2017don}, and the Sigmoid scheme from left to right.
The learning curves including the training loss (blue), validation loss (red), and validation accuracy (green) 
for VGG11 (middle) and ResNet18 (bottom) based on CIFAR10, are computed using each learning rate annealing with batch size $|\beta| = 16$ and momentum = 0.}
\label{fig:LRschedules.curves}
\end{figure*}
%

%
\def \pw {6mm}
\def \qw {6mm}
\begin{table*}[ht]
\caption{Validation accuracy ($\%$) of SGD using VGG11 (top) and ResNet18 (bottom) based on CIFAR10 with different batch size ($|\beta|$) and momentum ($m$). The average accuracy (upper) over the last 10$\%$ of epochs and the maximum (lower) are computed over 10 trials.}
\label{tab:SGD.BandMom}
\centering
%
%
\small
%
%
\setlength\extrarowheight{3pt}
\begin{tabular}{l | P{\pw}P{\pw}P{\pw}P{\qw} |  P{\pw}P{\pw}P{\pw}P{\qw} |  P{\pw}P{\pw}P{\pw}P{\qw} |  P{\pw}P{\pw}P{\pw}P{\qw}}
\hline
$\batchsize$  & 16 & 16 & 16 & 16 & 32 & 32 & 32 & 32 & 64 & 64 & 64 & 64 & 128 & 128 & 128 & 128 \\
$m$ & 0 & 0.3 & 0.6 & 0.9 & 0 & 0.3 & 0.6 & 0.9 & 0 & 0.3 & 0.6 & 0.9 & 0 & 0.3 & 0.6 & 0.9 \\
\hline
mean & 92.03 & 92.05 & 91.82 & 86.07 & 92.01 & 92.08 & 92.09 & 90.24 & 91.46 & 91.60 & 91.97 & 91.60 & 90.96 & 91.11 & 91.55 & 91.97\\
max & 92.48 & 92.33 & 92.45 & 87.43 & 92.43 & 92.56 & 92.48 & 90.99 & 91.93 & 91.92 & 92.39 & 92.01 & 91.30 & 91.47 & 91.81 & 92.34\\
\hline
mean & 94.89 & 94.89 & 94.50 & 90.77 & 94.72 & 94.90 & 94.85 & 92.64 & 94.14 & 94.36 & 94.71 & 94.14 & 93.33 & 93.75 & 94.20 & 94.49\\
max & 95.16 & 95.20 & 94.82 & 91.61 & 95.13 & 95.18 & 95.22 & 93.14 & 94.40 & 94.73 & 95.05 & 94.42 & 93.80 & 94.14 & 94.44 & 94.77\\
\hline
\end{tabular}\\
\vspace{2pt}
\end{table*}
%
%

%
%
%
%
%
%
%
%
\section{Experimental results} \label{sec:results}
We provide quantitative evaluation of our algorithm in comparison to the state-of-the-art optimization algorithms.
For the experiments, we use datasets including CIFAR10~\cite{krizhevsky2009learning} that consists of $50$K training and $10$K test object images with 10 categories, SVHN~\cite{netzer2011reading} that consists of $70$K training and $25$K test street scenes for digit recognition, and STL10~\cite{coates2011analysis} that consists of $500$ training and $800$ test object images with 10 categories.
Regarding the neural network architecture, we consider VGG11~\cite{simonyan2014very}, ResNet18~\cite{he2016deep,he2016identity}, and ResNet50~\cite{Balduzzi2017shattered} in combination with the batch normalization~\cite{ioffe2015batch}.
%
%
%
%
%
\par
In our comparative analysis, we consider the following optimization algorithms: SGD, Dropout (Drop)~\cite{srivastava2014dropout}, Adaptive Dropout (aDrop)~\cite{huang2016deep}, Entropy-SGD (eSGD)~\cite{Chaudhari2017EntropySGD}, Accelerated-SGD (aSGD)~\cite{Kidambi2018Acc}, and our proposed algorithm (Ours).
We use SGD as a baseline for comparing the performance of the above algorithms of which the common hyperparameters such as learning rate, batch size, and momentum are chosen with respect to the best validation accuracy of the baseline.
Regarding the additional parameters specific to Dropout and Entropy-SGD, we use the recommended values; the dropout rate is $0.5$ and the number of inner loop is $5$, respectively.
In the quantitative evaluation, we perform 10 independent trials and their average learning curves indicating the validation accuracy are considered. %
In particular, we consider both the maximum across all the epochs and the average over the last 10$\%$ epochs for the validation accuracy.
We present the numerical results that determine the hyperparameters used across all the experiments in the following section.
%
%
%
%
%
%
%
\subsection{Selection of optimal hyperparameters} \label{sec:hyper-parameters}
The learning rate, batch size, and momentum are selected by the best validation accuracy averaged over the last 10$\%$ of epochs and 10 trials by the baseline SGD using the models VGG11 and ResNet18 based on the dataset CIFAR10.
\par
\vspace{3pt}
\noindent {\bf Learning rate:}
In order to focus on the batch size, we carefully choose the learning rate as follow.  We have performed a comparative analysis using constant learning rates, staircase decay~\cite{smith2017don}, exponential decay~\cite{george2006adaptive}, sigmoid scheduling, RMSProp~\cite{tieleman2012lecture} and Adam~\cite{kingma2014adam} based on the networks VGG11 and ResNet18 using CIFAR10 dataset.
We used grid search, and set $0.1$ and $0.05$ as the initial value and exponential power, respectively for Exponential decay, $0.95$ as the weighting factor for RMSProp, $0.9$ and $0.999$ as the weighting factors of gradients and their moments, respectively for Adam.  We used $0.1$ and $0.001$ as the initial and final values with the steepness parameter for the sigmoid annealing.
\par
%
The quantitative evaluation in validation accuracy for each learning rate scheme is presented in Table~\ref{tab:LRschedules.accuracy} where the average over the last 10$\%$ of epochs (upper) and the maximum (lower) are obtained by 10 trials. 
The learning rates by selected schemes are presented in Figure~\ref{fig:LRschedules.curves} (top) and the learning curves including the training loss (blue), validation loss (red), and validation accuracy (green) are presented in Figure~\ref{fig:LRschedules.curves} using VGG11 (middle) and ResNet18 (bottom).
The sigmoid scheduling has been shown to achieve the best validation accuracy while yielding smooth validation curves, thus we apply the sigmoid learning rate to all the optimization algorithms throughout the following experiments.

\par
\vspace{3pt}
\noindent {\bf Batch size and momentum:} 
The selection of batch size is related to the momentum coefficient and we search for their optimal combination in terms of the validation accuracy by SGD using VGG11 and ResNet18 based on CIFAR10. 
We use $16, 32, 64, 128$ for batch size and $0, 0.3, 0.6, 0.9$ for momentum, respectively. The validation accuracy in average and maximum is obtained for VGG11 (upper) and ResNet18 (lower) with each combination of the parameters in Table~\ref{tab:SGD.BandMom} indicating that better results are obtained by the combination of larger both batch size and momentum, or smaller both batch size and momentum leading to our choice of momentum $0, 0.3, 0.6, 0.9$ for batch size $16, 32, 64, 128$, respectively throughout the experiments.
%
%

%
%
\def \pw {4.8mm}
\def \qw {6.2mm}
\begin{table*} [ht]
\caption{Validation accuracy ($\%$) of our algorithm is computed with varying parameters for (1) the local ($\alpha$) and  (2) the global ($\paramGlobal$) adaptive probabilities using VGG11 (left) and ResNet18 (right) based on CIFAR10 with batch size $16$.
The average accuracy (upper) over the last 10$\%$ epochs and the maximum (lower) over 10 trials are shown.}
\label{tab:analysis.parameter}
\centering
\small
\hspace*{-14pt} 
\setlength\extrarowheight{3pt}
\begin{tabular}{cc}
	(1) Effect of the local parameter ($\alpha$) with $\paramGlobal = 0$ & (2) Effect of the global parameter ($\paramGlobal$) with $\alpha = 0.1$\\
	
	\begin{tabular}{P{6mm} | P{5.2mm}P{\pw}P{\pw}P{\qw} | P{5.2mm}P{\pw}P{\pw}P{\qw} }
	\hline
	 & \multicolumn{4}{c|}{VGG11} &\multicolumn{4}{c}{ResNet18} \\
	\cline{2-9}
	$\alpha$ & -0.1 & 0 & 0.1 & 1 & -0.1 & 0 & 0.1 & 1\\
	\hline
	mean & 92.09 & 92.17 & \textbf{92.18} & 91.78 & 94.85 & 94.86 & \textbf{94.95} & 94.79\\
	max & 92.44 & 92.48 & \textbf{92.52} & 92.28 & 95.13 & 95.11 & \textbf{95.27} & 95.08\\
	\hline
	\end{tabular} 
	&
	\begin{tabular}{P{6mm} | P{\pw}P{\pw}P{\pw}P{\qw} | P{\pw}P{\pw}P{\pw}P{\qw} }
	\hline
	 & \multicolumn{4}{c|}{VGG11} &\multicolumn{4}{c}{ResNet18} \\
	\cline{2-9}
	$\paramGlobal$ & 0 & -2 & -4 & -8 & 0 & -2 & -4 & -8\\
	\hline
	mean & 92.18 & 92.35 & 92.38 & \textbf{92.46} & 94.95 & 95.06 & 95.05 & \textbf{95.15}\\
	max & 92.52 & 92.85 & \textbf{92.88} & 92.74 & 95.27 & 95.40 & \textbf{95.60} & 95.46\\
	\hline
	\end{tabular}\\
\end{tabular}
\vspace{2pt}
\end{table*}
%
%
%
%
%
%
%
%

\def \pw {12mm}
\def \qw {12mm}
\begin{table*}[htb]
\caption{Validation accuracy ($\%$) is computed with the algorithms: Stochastic gradient descent (SGD) and our algorithm (Ours) with the models: VGG11 (top) and ResNet18 (bottom) based on the datasets: CIFAR10 using its partial training data: $1/2$, $1/4$, $1/8$ and the batch sizes $|\beta|$: 16, 32, 64, 128. The average validation accuracy over the last 10$\%$ of epochs is computed over 10 trials. The accuracy gain by our algorithm with respect to SGD is also presented for ease of interpretation.}
\label{tab:partial_training.v3}
%
\centering
\small
%
%
\setlength\extrarowheight{3pt}
\begin{tabular}{l | P{\pw}P{\pw}P{\qw} | P{\pw}P{\pw}P{\qw} | P{\pw}P{\pw}P{\qw}}
\multicolumn{10}{c}{\small (1) Validation accuracy ($\%$) and improvement ($\%$-point) for VGG11}  \\
\hline
 & \multicolumn{3}{c|}{SGD}  & \multicolumn{3}{c|}{Ours}  & \multicolumn{3}{c}{improvement}  \\
 \hline
training ratio & $1/2$ & $1/4$ & $1/8$ & $1/2$ & $1/4$ & $1/8$ & $1/2$ & $1/4$ & $1/8$ \\
\hline
$\batchsize$=16 & 89.16 & 84.78 & 79.25 & 89.58 & 85.50 & 80.18 & 0.43 & 0.71 & {0.93} \\
$\batchsize$=32 & 88.87 & 84.61 & 79.27 & 89.23 & 85.22 & 79.64 & 0.36 & {0.61} & 0.37 \\
$\batchsize$=64 & 88.79 & 84.43 & 78.44 & 89.08 & 84.81 & 79.50 & 0.29 & 0.37 & {1.05} \\
$\batchsize$=128 & 88.70 & 84.07 & 77.47 & 89.02 & 84.73 & 78.98 & 0.33 & 0.67 & {1.51} \\
\hline
\end{tabular}\\
 \vskip 0.12in
\begin{tabular}{l | P{\pw}P{\pw}P{\qw} | P{\pw}P{\pw}P{\qw} | P{\pw}P{\pw}P{\qw}}
\multicolumn{10}{c}{\small (2) Validation accuracy ($\%$) and improvement ($\%$-point) for ResNet18}  \\
\hline
 & \multicolumn{3}{c|}{SGD}  & \multicolumn{3}{c|}{Ours}  & \multicolumn{3}{c}{improvement}  \\
 \hline
training ratio & $1/2$ & $1/4$ & $1/8$ & $1/2$ & $1/4$ & $1/8$ & $1/2$ & $1/4$ & $1/8$ \\
\hline
$\batchsize$=16 & 92.32 & 88.34 & 82.77 & 92.53 & 88.63 & 83.68 & 0.22 & 0.29 & {0.92}\\
$\batchsize$=32 & 92.01 & 87.88 & 81.92 & 92.31 & 88.37 & 82.53 & 0.29 & 0.49 & {0.60}\\
$\batchsize$=64 & 91.81 & 87.42 & 81.50 & 92.23 & 87.73 & 82.02 & 0.43 & 0.30 & {0.52}\\
$\batchsize$=128 & 91.84 & 87.43 & 81.02 & 92.09 & 87.96 & 81.64 & 0.25 & 0.53 & {0.62}\\
\hline
\end{tabular}\\
\end{table*}
%
%
%
%
%
%
%

\graphicspath{{./figure/img/}}
\def \fw {130pt}   
\def \pw {120pt}   
\begin{figure*} [htb]
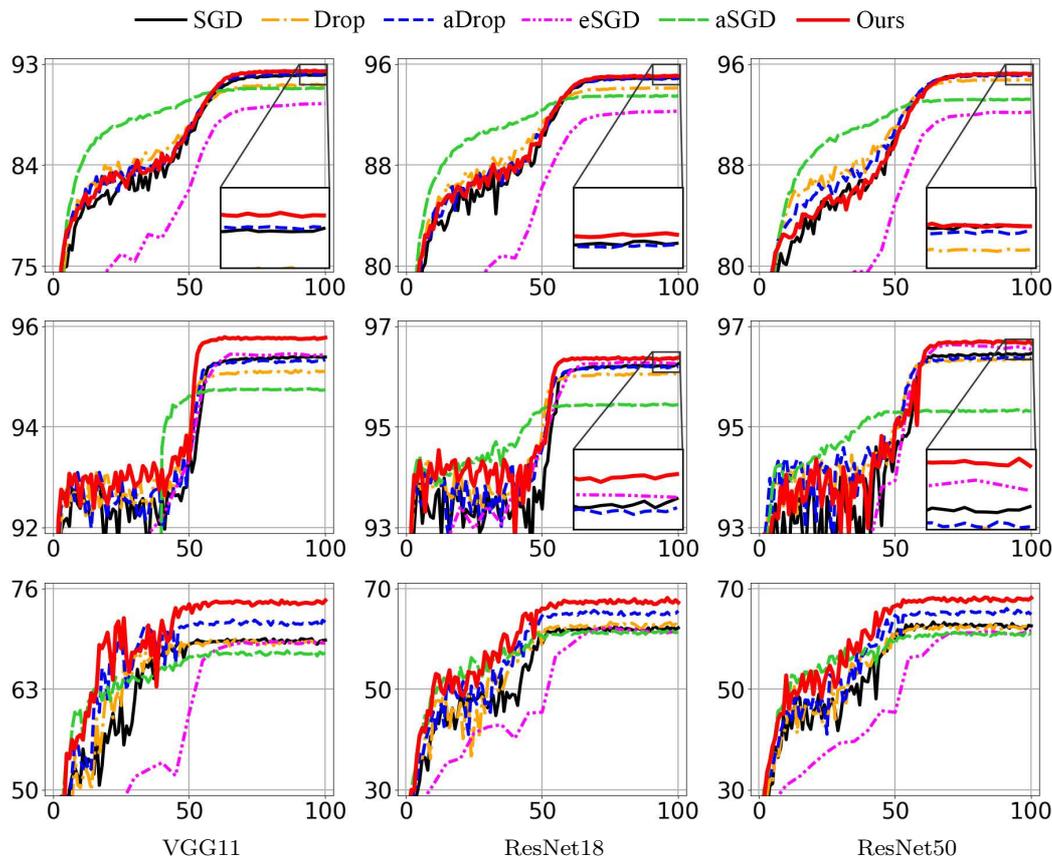

\small
\centering
\includegraphics[width=0.60\textwidth]{{{legend2.2}}} \\
\begin{tabular}{P{\pw}P{\pw}P{\pw}} 
\includegraphics[width=\fw]{{{cmpAccuracy,cifar10,VGG11,e,100,B,16,mo,0.0}}} &
\includegraphics[width=\fw]{{{cmpAccuracy,cifar10,ResNet18,e,100,B,16,mo,0.0}}} &
\includegraphics[width=\fw]{{{cmpAccuracy,cifar10,ResNet50,e,100,B,16,mo,0.0}}} \\
\includegraphics[width=\fw]{{{cmpAccuracy,SVHN,VGG11,e,100,B,16,mo,0.0}}} &
\includegraphics[width=\fw]{{{cmpAccuracy,SVHN,ResNet18,e,100,B,16,mo,0.0}}} &
\includegraphics[width=\fw]{{{cmpAccuracy,SVHN,ResNet50,e,100,B,16,mo,0.0}}} \\
\includegraphics[width=\fw]{{{cmpAccuracy,STL10,VGG11_STL10,e,100,B,16,mo,0.0}}} &
\includegraphics[width=\fw]{{{cmpAccuracy,STL10,ResNet18_STL10,e,100,B,16,mo,0.0}}} &
\includegraphics[width=\fw]{{{cmpAccuracy,STL10,ResNet50_STL10,e,100,B,16,mo,0.0}}} \\
\quad\quad\quad VGG11 & \quad\quad\quad ResNet18 & \quad\quad\quad ResNet50\\
\end{tabular} \\
%
%
\vskip 0.1in
\caption{Validation accuracy curve for the datasets: CIFAR10 (top), SVHN (middle), STL10 (bottom)
by the models: VGG11 (left), ResNet18 (middle), ResNet50 (right) optimized using SGD (black), Dropout (yellow), Adaptive Dropout (blue), Entropy-SGD (magenta), Accelerated-SGD (green), and Ours (red). The x-axis represents epoch and the y-axis represents accuracy ($\%$).}
\label{fig:comparison_with_others.accuracy_curves}
\end{figure*}
%

%
%
%
%
%
%
\subsection{Effect of local and global adaptive probability}
We empirically demonstrate the effect of parameter $\alpha$ for the local adaptive probability in Eq.~\eqref{eq:prob:local} and parameter $\lambda$ for the global one in Eq.~\eqref{eq:prob:global} in our algorithm with varying one parameter while the other is fixed. 
We present the average and maximum validation accuracy in Table~\ref{tab:analysis.parameter} where (1) the parameter $\alpha$ in local adaptive probability is set as $-0.1, 0, 0.1, 1$ with fixed $\lambda = 0$, and (2) the parameter $\lambda$ in global adaptive probability is set as $0, -2, -4, -8$ with fixed $\alpha = 0.1$, respectively using VGG11 (left) and ResNet18 (right) based on CIFAR10 with the batch size being 16.
It is shown that $\alpha = 0.1$ and $\lambda = -4$ yields the best results for VGG11 and ResNet18, and thus we use the pair of $\alpha = 0.1$ and $\lambda = -4$ for our combined adaptive scheme throughout the experiments.
%
%
%

%
%
%
\subsection{Effect of stochastic batch size in generalization}
We empirically demonstrate that the proposed algorithm with the combined adaptive scheme achieves better generalization than the baseline SGD in terms of the validation accuracy obtained with partial training data. 
In Table~\ref{tab:partial_training.v3}, we present the average validation accuracy with VGG11 (top) and ResNet18 (bottom) using CIFAR10 
from which $1/2, 1/4, 1/8$ portions of the 50K training data are randomly selected for each individual trial and used its training process.
We use $16, 32, 64, 128$ as universal batch sizes for each partial set of data.
It is clearly shown that our algorithm yields better accuracy than SGD irrespective of the batch size.
Moreover the performance gain by our algorithm increases as less number of training data are used due to our effective generalization imposed by adaptive regularization with the local and global update probabilities.
%
%

%
%
%
%
%
%
%
%
%
%
%
%
\subsection{Comparison to the state-of-the-art}
We now compare our algorithm with the combined adaptive probability to SGD, Dropout (Drop)~\cite{srivastava2014dropout}, Adaptive Dropout (aDrop)~\cite{huang2016deep}, Entropy-SGD (eSGD)~\cite{Chaudhari2017EntropySGD}, and Accelerated-SGD (aSGD)~\cite{Kidambi2018Acc}.
We present the average (upper) and maximum (lower) validation accuracy with the models VGG11 (left), ResNet18 (middle), ResNet50 (right) based on the datasets (1) CIFAR10, (2) SVHN, and (3) STL10 in Table~\ref{tab:comparison_with_others.accuracy} where 16, 32, 64, 128 are used for the universal batch size $\batchsize$.
It is shown that our algorithm outperforms the others consistently regardless of model, dataset and batch size.
Figure~\ref{fig:comparison_with_others.accuracy_curves} also visualizes validation accuracy curves with batch size of 16, indicating that ours (red) achieves better accuracy than SGD (black), Dropout (yellow), Adaptive Dropout (blue), Entropy-SGD (magenta), and Accelerated-SGD (green) across all the models VGG11 (left), ResNet18 (middle), ResNet50 (right) and all the datasets CIFAR10 (top), SVHN (middle), and STL10 (bottom).
%
%
%
%
%

%
%
%
%
\section{Conclusion and discussion} \label{sec:conculusion}
We have proposed a first-order optimization algorithm with stochastic batch size for large scale problems in deep learning. 
Our algorithm determines batch size for each parameter at each iteration in a stochastic way following a probability proportional to its gradient norm in such a way that adaptive regularization is imposed on each parameter, leading to better generalization of the network model.
The efficient computation of the gradient with varying batch size is achieved by the cumulative moving average scheme based on the usual stochastic gradients.
The effectiveness of the proposed algorithm has been demonstrated by the experimental results in which our algorithm outperforms a number of other methods for the classification task with conventional network architectures using a number of benchmark datasets indicating that our algorithm achieves better generalization in particular with less number of training data without additional computational cost.
%
%
%

\begin{acknowledgements}
This work was supported by the National Research Foundation of Korea: NRF-2017R1A2B4006023 and NRF-2018R1A4A1059731.
\end{acknowledgements}

\def \ow{4.2mm}
\def \pw{4.2mm}
\def \qw{6.5mm}
\begin{table*}[!h]
\caption{Validation accuracy ($\%$) is computed with the algorithms: Stochastic gradient descent (SGD), Dropout (Drop)~\cite{srivastava2014dropout}, Adaptive Dropout (aDrop)~\cite{huang2016deep}, Entropy-SGD (eSGD)~\cite{Chaudhari2017EntropySGD}, Accelerated-SGD (aSGD)~\cite{Kidambi2018Acc}, and our algorithm (Ours) with the models: VGG11 (left), ResNet18 (middle), ResNet50 (right) based on the datasets: CIFAR10 (top block), SVHN (middle), and STL10 (bottom) using the batch sizes $|\beta|$: 16, 32, 64, 128. The average validation accuracy over the last 10$\%$ of epochs is shown at the upper part of each block and the maximum validation accuracy over all the epochs is shown at the lower part. The accuracy is computed over 10 trials.}
\label{tab:comparison_with_others.accuracy}
\vskip 0.15in
\centering
\setlength\extrarowheight{5pt}
\scriptsize
{\small (1) Validation accuracy for CIFAR10 with the average over the last 10$\%$ epochs (upper part) and the maximum (lower part)}\\
\begin{tabular}{| l | P{\ow}P{\ow}P{\pw}P{\pw}P{\pw}P{\qw} |  P{\ow}P{\ow}P{\pw}P{\pw}P{\pw}P{\qw} | P{\ow}P{\ow}P{\pw}P{\pw}P{\pw}P{\qw} |}
\hline
& \multicolumn{6}{c|}{VGG11}  & \multicolumn{6}{c|}{ResNet18}  & \multicolumn{6}{c |}{ResNet50} \\
\cline{2-19}
mean & SGD & Drop & aDrop & eSGD & aSGD & Ours & SGD & Drop & aDrop & eSGD & aSGD & Ours & SGD & Drop & aDrop & eSGD & aSGD & Ours\\
\hline
$\batchsize$=16 & 92.03 & 91.17 & 92.10 & 89.47 & 90.86 & \textbf{92.38} & 94.89 & 94.11 & 94.84 & 92.26 & 93.49 & \textbf{95.05} & 95.22 & 94.76 & 95.11 & 92.18 & 93.21 & \textbf{95.24}\\
$\batchsize$=32 & 92.08 & 90.94 & 92.00 & 89.38 & 90.83 & \textbf{92.36} & 94.90 & 94.04 & 94.79 & 91.97 & 93.42 & \textbf{94.99} & 95.20 & 94.62 & 95.01 & 91.74 & 93.03 & \textbf{95.29}\\
$\batchsize$=64 & 91.97 & 90.73 & 91.88 & 89.12 & 90.72 & \textbf{92.24} & 94.71 & 93.76 & 94.63 & 91.00 & 93.11 & \textbf{94.92} & 94.93 & 94.39 & 94.91 & 90.47 & 92.84 & \textbf{95.19}\\
$\batchsize$=128 & 91.97 & 90.48 & 91.88 & 88.83 & 90.64 & \textbf{92.34} & 94.49 & 93.85 & 94.58 & 91.48 & 92.98 & \textbf{94.92} & 94.48 & 94.37 & 94.54 & 90.43 & 92.71 & \textbf{94.81}\\
\hline
max  & SGD & Drop & aDrop & eSGD & aSGD & Ours & SGD & Drop & aDrop & eSGD & aSGD & Ours & SGD & Drop & aDrop & eSGD & aSGD & Ours\\
\hline
$\batchsize$=16 & 92.48 & 91.50 & 92.34 & 89.81 & 91.27 & \textbf{92.88} & 95.16 & 94.42 & 95.12 & 92.49 & 93.86 & \textbf{95.60} & 95.53 & 95.08 & 95.41 & 92.56 & 93.82 & \textbf{95.58}\\
$\batchsize$=32 & 92.56 & 91.38 & 92.37 & 89.77 & 91.38 & \textbf{92.62} & 95.18 & 94.36 & 95.13 & 92.29 & 93.88 & \textbf{95.32} & 95.54 & 94.94 & 95.27 & 92.42 & 93.45 & \textbf{95.67}\\
$\batchsize$=64 & 92.39 & 91.23 & 92.31 & 89.59 & 91.24 & \textbf{92.70} & 95.05 & 94.05 & 95.05 & 92.08 & 93.62 & \textbf{95.33} & 95.41 & 94.84 & 95.28 & 90.90 & 93.51 & \textbf{95.45}\\
$\batchsize$=128 & 92.34 & 91.20 & 92.25 & 89.20 & 91.08 & \textbf{92.84} & 94.77 & 94.07 & 94.78 & 91.69 & 93.34 & \textbf{95.15} & 94.98 & 94.66 & 94.91 & 91.28 & 93.30 & \textbf{95.14}\\
\hline
\end{tabular} \\
\vskip 0.1in
{\small (2) Validation accuracy for SVHN with the average over the last 10$\%$ epochs (upper part) and the maximum (lower part)}\\
\begin{tabular}{| l | P{\ow}P{\ow}P{\pw}P{\pw}P{\pw}P{\qw} |  P{\ow}P{\ow}P{\pw}P{\pw}P{\pw}P{\qw} | P{\ow}P{\ow}P{\pw}P{\pw}P{\pw}P{\qw} |}
\hline
& \multicolumn{6}{c|}{VGG11}  & \multicolumn{6}{c|}{ResNet18}  & \multicolumn{6}{c |}{ResNet50} \\
\cline{2-19}
mean & SGD & Drop & aDrop & eSGD & aSGD & Ours & SGD & Drop & aDrop & eSGD & aSGD & Ours & SGD & Drop & aDrop & eSGD & aSGD & Ours\\
\hline
$\batchsize$=16 & 95.39 & 95.11 & 95.32 & 95.43 & 94.75 & \textbf{95.76} & 96.22 & 96.06 & 96.19 & 96.27 & 95.44 & \textbf{96.36} & 96.45 & 96.34 & 96.37 & 96.57 & 95.32 & \textbf{96.68}\\
$\batchsize$=32 & 95.42 & 95.07 & 95.35 & 95.38 & 94.77 & \textbf{95.72} & 96.17 & 95.98 & 96.19 & 96.13 & 95.57 & \textbf{96.30} & 96.41 & 96.23 & 96.29 & 96.33 & 95.38 & \textbf{96.59}\\
$\batchsize$=64 & 95.37 & 95.08 & 95.33 & 95.36 & 94.78 & \textbf{95.72} & 96.08 & 95.90 & 96.12 & 95.74 & 95.49 & \textbf{96.28} & 96.31 & 96.17 & 96.26 & 96.16 & 95.36 & \textbf{96.57}\\
$\batchsize$=128 & 95.47 & 95.13 & 95.35 & 95.20 & 94.62 & \textbf{95.74} & 96.16 & 95.94 & 96.10 & 95.85 & 95.35 & \textbf{96.33} & 96.41 & 96.29 & 96.31 & 95.90 & 94.86 & \textbf{96.63}\\
\hline
max  & SGD & Drop & aDrop & eSGD & aSGD & Ours & SGD & Drop & aDrop & eSGD & aSGD & Ours & SGD & Drop & aDrop & eSGD & aSGD & Ours\\
\hline
$\batchsize$=16 & 95.63 & 95.24 & 95.44 & 95.60 & 95.01 & \textbf{95.92} & 96.38 & 96.22 & 96.35 & 96.42 & 95.67 & \textbf{96.48} & 96.61 & 96.54 & 96.51 & 96.75 & 95.66 & \textbf{96.88}\\
$\batchsize$=32 & 95.59 & 95.24 & 95.59 & 95.57 & 95.03 & \textbf{95.87} & 96.42 & 96.16 & 96.34 & 96.24 & 95.74 & \textbf{96.50} & 96.56 & 96.39 & 96.49 & 96.64 & 95.72 & \textbf{96.82}\\
$\batchsize$=64 & 95.58 & 95.31 & 95.51 & 95.63 & 95.00 & \textbf{95.91} & 96.25 & 96.04 & 96.28 & 96.10 & 95.77 & \textbf{96.47} & 96.54 & 96.37 & 96.42 & 96.45 & 95.64 & \textbf{96.86}\\
$\batchsize$=128 & 95.59 & 95.34 & 95.57 & 95.41 & 94.91 & \textbf{95.91} & 96.26 & 96.10 & 96.29 & 95.98 & 95.55 & \textbf{96.49} & 96.75 & 96.55 & 96.56 & 96.13 & 95.42 & \textbf{96.85}\\
\hline
\end{tabular} \\
\vskip 0.1in
{\small (3) Validation accuracy for STL10 with the average over the last 10$\%$ epochs (upper part) and the maximum (lower part)}\\
\begin{tabular}{| l | P{\ow}P{\ow}P{\pw}P{\pw}P{\pw}P{\qw} |  P{\ow}P{\ow}P{\pw}P{\pw}P{\pw}P{\qw} | P{\ow}P{\ow}P{\pw}P{\pw}P{\pw}P{\qw} |}
\hline
& \multicolumn{6}{c|}{VGG11}  & \multicolumn{6}{c|}{ResNet18}  & \multicolumn{6}{c |}{ResNet50} \\
\cline{2-19}
mean & SGD & Drop & aDrop & eSGD & aSGD & Ours & SGD & Drop & aDrop & eSGD & aSGD & Ours & SGD & Drop & aDrop & eSGD & aSGD & Ours\\
\hline
$\batchsize$=16 & 69.31 & 69.11 & 71.53 & 69.03 & 67.62 & \textbf{74.21} & 62.14 & 62.84 & 65.09 & 61.38 & 61.41 & \textbf{67.42} & 62.47 & 62.33 & 65.30 & 61.31 & 60.96 & \textbf{67.88}\\
$\batchsize$=32 & 69.62 & 69.43 & 71.76 & 67.89 & 67.01 & \textbf{74.34} & 63.12 & 62.34 & 64.97 & 61.52 & 62.51 & \textbf{66.16} & 63.16 & 61.86 & 64.17 & 61.55 & 62.94 & \textbf{66.60}\\
$\batchsize$=64 & 68.29 & 68.02 & 71.24 & 66.86 & 66.57 & \textbf{73.05} & 62.65 & 61.28 & 62.91 & 55.39 & 62.43 & \textbf{64.74} & 62.31 & 61.56 & 63.70 & 54.03 & 61.89 & \textbf{65.06}\\
$\batchsize$=28 & 67.23 & 66.23 & 67.58 & 65.04 & 65.43 & \textbf{70.00} & 61.28 & 61.11 & 62.93 & 48.53 & 60.15 & \textbf{63.38} & 60.61 & 62.00 & 61.40 & 48.84 & 61.27 & \textbf{64.56}\\
\hline
max  & SGD & Drop & aDrop & eSGD & aSGD & Ours & SGD & Drop & aDrop & eSGD & aSGD & Ours & SGD & Drop & aDrop & eSGD & aSGD & Ours\\
\hline
$\batchsize$=16 & 71.18 & 73.21 & 73.34 & 70.75 & 69.54 & \textbf{75.79} & 66.15 & 66.49 & 69.23 & 65.91 & 64.93 & \textbf{70.35} & 65.58 & 66.09 & 68.28 & 63.29 & 65.54 & \textbf{71.01}\\
$\batchsize$=32 & 72.26 & 71.43 & 73.63 & 71.33 & 69.43 & \textbf{76.26} & 65.89 & 65.15 & 68.44 & 64.80 & 67.08 & \textbf{68.94} & 66.50 & 65.74 & 67.38 & 65.00 & 67.94 & \textbf{69.29}\\
$\batchsize$=64 & 70.79 & 70.91 & 73.38 & 69.04 & 68.95 & \textbf{74.63} & 65.88 & 65.80 & 67.48 & 60.35 & 66.06 & \textbf{67.84} & 65.69 & 65.26 & 65.98 & 62.01 & 64.35 & \textbf{66.93}\\
$\batchsize$=128 & 69.68 & 68.44 & 69.98 & 66.69 & 68.74 & \textbf{71.86} & 65.30 & 64.69 & 66.71 & 55.26 & 65.21 & \textbf{66.99} & 66.01 & 65.95 & 64.93 & 55.55 & 64.21 & \textbf{67.54}\\
\hline
\end{tabular} \\
\vspace{80pt}
\end{table*}
%

\bibliographystyle{spmpsci}      
\bibliography{mybib}   

\end{document}